# PATH PLANNING FOR AUTONOMOUS VEHICLES WITH MINIMAL COLLISION SEVERITY

Qiannan Wang[1][a], Matthias Gerdts[1][b]

[1]Institute of Applied Mathematics and Scientific Computing, Universität der Bundeswehr München, Germany
{qiannan.wang, matthias.gerdts}@unibw.de

Keywords: path planning, collision severity, dynamic obstacles, autonomous driving

Abstract: This paper proposes a path planning algorithm for autonomous vehicles, evaluating collision severity with respect to both static and dynamic obstacles. A collision severity map is generated from ratings, quantifying the severity of collisions. A two-level optimal control problem is designed. At the first level, the objective is to identify paths with the lowest collision severity. Subsequently, at the second level, among the paths with lowest collision severity, the one requiring the minimum steering effort is determined. Finally, numerical simulations were conducted using the optimal control software OCPID-DAE1. The study focuses on scenarios where collisions are unavoidable. Results demonstrate the effectiveness and significance of this approach in finding a path with minimum collision severity for autonomous vehicles. Furthermore, this paper illustrates how the ratings for collision severity influence the behaviour of the automated vehicle.

[a] https://orcid.org/0000-0000-0000-0000
[b] https://orcid.org/0000-0000-0000-0000

# 1 INTRODUCTION

Currently, autonomous driving is compelling and trending for both manufacturers and researchers. It is regarded as a new trend and challenge for the automobile industry. Autonomous driving is categorized into several levels, from "no automation" to "full automation". To achieve fully automated driving, further concentration should be focused on urban areas, where the environment is more complicated and safety problems are more urgent.

Path planning is a necessary step for autonomous driving. It acts as the "brain" of the vehicle by defining the way to the destination while simultaneously satisfying constraints. It has been widely studied, and a large number of methods have been developed. Some popular methods are: (1) graph search-based methods, such as Dijkstra's algorithm introduced in [1,2], and the A* method developed in [3]; (2) sampling-based methods, e.g. Probabilistic Roadmap Method (PRM) [4] and Rapidly-exploring Random Tree (RRT) [5, 6]; (3) interpolating curve based methods, e.g. spline curves [7], polynomial curves [8,9]; (4) numerical optimization methods, e.g. model predictive control (MPC) [10,11]. Generally, graph search-based and sampling-based methods are suitable for finding a global path. Paths generated by interpolating curve-based methods are usually not continuous. Optimization-based methods are available for local path planning and are capable of taking constraints into account. In this paper, we choose a numerical optimization method.

When autonomous vehicles operate in uncertain or populated environments, it is essential to account for risks in order to generate an effective and safe path. Research on path planning based on risk awareness has been conducted by several researchers for unmanned aerial vehicles [12, 13, 14, 15, 16]. The general concept is to compute an effective path that minimizes the risk to the population and its surroundings. In [16], a B-spline method is employed to model a risk map based on various objects. However, this approach only considers static obstacles. In [13], a risk assessment technology and bi-objective optimization methods are developed to find a low-risk and efficient solution. The risk value of an object is defined on a scale between 0 and 1. Nevertheless, in typical scenarios, autonomous vehicles encounter various objects in complex environments, including pedestrians, bicyclists, trees, and other vehicles. Assigning the same risk value for all types of participants is not sufficiently informative. In order to generate a collision-free path or reduce the severity of collisions, distinct risk values for different objects need to be precisely defined. Therefore, this paper discusses a new approach for modelling collision severity regarding dynamic obstacles and optimization for path planning.

To achieve fully autonomous driving in urban environments, autonomous vehicles should have the ability of avoiding collisions, whenever possible, with any other objects in the area. When time to collision is within 1-2s, existing solutions such as emergency braking is not efficient to avoid collision. Therefore, assessing the collision severity is meaningful to minimize the harm to the greatest extent. To address this challenge, a minimum collision severity path planning method is investigated. The main contributions of this work include:

(1) The use of a mathematical method to model the collision severities posed by both static and dynamic obstacles. Here, all encountered objects are classified, and corresponding collision severity values are precisely defined.

(2) The design of a two-level optimal control problem based on the collision severity map for path planning, which considers both minimizing collision severity and reducing steering efforts. Herein, the minimization of collision severity has higher priority.

The paper's structure is as follows. The method of generating collision severity is described in section 2. Detailed information about the path planning method is presented in section 3. Simulation results are discussed in section 4, and conclusions and future work are provided in section 5.

# 2 DEFINITION OF COLLISION SEVERITY

To generate the safest possible path, a well-defined collision severity map for each specific scenario is essential. In the context of autonomous driving in an urban environment, one of the biggest challenges is handling moving objects in the area. Therefore, precisely describing the collision severity posed by dynamic objects is of great significance. In this section, we introduce a mathematical method to assess the collision severity, with respect to both static and dynamic obstacles. Furthermore, we precisely define collision severities for various objects.

*Remarks: In this paper, for simplicity, we restrict the discussion to collision severities that depend on the position (x,y) of the objects and the relative velocities between the object and the ego vehicle in the traffic scenario. However, it is straightforward, but technically more involved, to consider collision severities that take into account additional factors such as collision angles and the position of the object in the collision (e.g. front, rear, or side).*

## 2.1 Collision Severity Model

In the past decades, one popular method for obstacle avoidance in robotics is the potential field method [17]. The basic concept of the potential field method is to fill the robot's workspace with an artificial potential field, in which the robot is attracted to its goal destination and repulsed away from obstacles. We adopt this idea to define the shapes of obstacles and assign corresponding collision severity values to them. The aim is to keep the autonomous vehicle away from the obstacles as far as possible. In this paper, two different geometric objects are defined, which can represent various objects that appear in the environment. One is a circular object for participants, such as pedestrians, and the other is a rectangular object for participants, such as cars, buses or construction sites. More complicated shapes of obstacles could be easily designed by overlapping these basic shape functions.

The circular object is defined by equation (1) and it is shown in Figure 1.

$$f_{dc}(x,y) = \begin{cases} 1, & \text{if } \|(x,y)\|_2 \leq 1 \\ \exp\left(-\left(\frac{\|(x,y)\|_2 - 1}{d}\right)^4\right), & otherwiese \end{cases} \quad (1)$$

The function is twice differentiable and could also be interpreted as a probability for the presence of the obstacle. The value d>0 can be used to model a fuzzy region around the obstacle where the probability of the presence of the obstacle decreases. In such a way uncertainties in measurements could be modelled. To describe various shapes of objects, a general function of an ellipse with radii a and b is given as follows:

$$f_{dc,a,b}(x,y) = f_{dc}\left(\frac{x}{a}, \frac{y}{b}\right). \quad (2)$$

The rectangular object is defined by equation (3) and it is shown in Figure 2. The function is differentiable.

$$f_{dr}(x,y) = \begin{cases} 1, & \text{if } \|(x,y)\|_\infty \leq 1, \\ \exp\left(-\left(\frac{\|(x,y)\|_\infty - 1}{d}\right)^4\right), \\ \quad \text{if } \|(x,y)\|_\infty > 1 \text{ and } |x| \leq 1 \text{ or } |y| \leq 1, \\ \exp\left(-\left(\frac{\|(x-1,y-1)\|_2}{d}\right)^4\right), \\ \quad \text{if } \|(x,y)\|_2 > \sqrt{2} \text{ and } x > 1 \text{ and } y > 1 \\ \exp\left(-\left(\frac{\|(x+1,y-1)\|_2}{d}\right)^4\right), \\ \quad \text{if } \|(x,y)\|_2 > \sqrt{2} \text{ and } x < -1 \text{ and } y > 1 \\ \exp\left(-\left(\frac{\|(x+1,y+1)\|_2}{d}\right)^4\right), \\ \quad \text{if } \|(x,y)\|_2 > \sqrt{2} \text{ and } x < -1 \text{ and } y < -1 \\ \exp\left(-\left(\frac{\|(x-1,y+1)\|_2}{d}\right)^4\right), \\ \quad \text{if } \|(x,y)\|_2 > \sqrt{2} \text{ and } x > 1 \text{ and } y < -1 \end{cases} \quad (3)$$

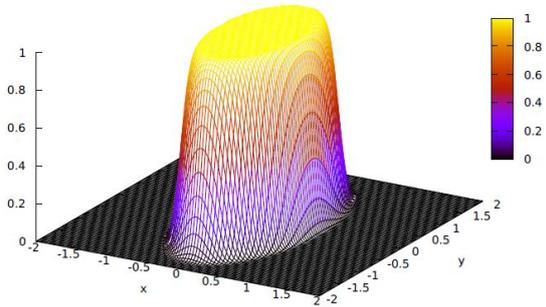

Figure 1: Circular object model.

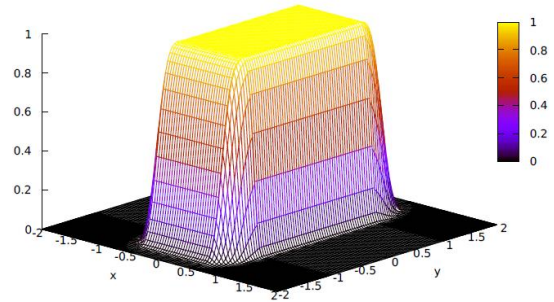

Figure 2: Rectangular object model.

A rectangular object of length a and width b is given by

$$f_{dr,a,b}(x,y) = f_{dr}(\frac{x}{a}, \frac{y}{b}) \quad (4)$$

The above functions are normalized in the sense that the centre is located at the origin and the orientation is aligned with the x-axis. In order to move the object to a possibly time-dependent position ($x_c(t)$, $y_c(t)$) and to rotate it by the angle $\varphi_c(t)$ around the vertical axis, we employ the coordinate transformation

$$\begin{pmatrix} x_p(t) \\ y_p(t) \end{pmatrix} = \begin{pmatrix} cos\varphi_c(t) & -sin\varphi_c(t) \\ sin\varphi_c(t) & cos\varphi_c(t) \end{pmatrix}^T \begin{pmatrix} x - x_c(t) \\ y - y_c(t) \end{pmatrix}$$
$$= \begin{pmatrix} (x - x_c(t))cos\varphi_c(t) + (y - y_c(t))sin\varphi_c(t) \\ -(x - x_c(t))sin\varphi_c(t) + (y - y_c(t))cos\varphi_c(t) \end{pmatrix} \quad (5)$$

Using this transformation, the collision severity for the obstacle with velocity $v_o(t)$ at time t with an object located at a point (x,y) with velocity v is modelled as

$$cs(t,x,y,v) = C * V_r(t) * f_d(x_p(t), y_p(t)) \quad (6)$$

Here, C is a constant collision severity value determined by the type of the object. $V_r(t)=v-v_o(t)$ is the relative velocity between the object and the ego vehicle. $f_d$ represents either $f_{dc}$, $f_{dr}$, $f_{dc,a,b}$ or $f_{dr,a,b}$. Detailed information on how to assign collision severity values for various objects will be provided in section 2.2.

## 2.2 Assignment Collision Severity Values

In this work, it is assumed that information of objects such as type of participants and location are already known through environment perception system. It will not be discussed here. To generate a meaningful collision severity map, assigning appropriate collision severity values to objects appearing in the scenario is crucial. It should be noted that the collision severity values determine the risk-minimal path and, consequently, the behaviour of the autonomous vehicle. Hence, assigning the collision severity values is a crucial task, particularly in situations where collisions are unavoidable. Precise collision severity values should be evaluated by an expert commission based on ethical and legal decisions and agreements. The assignment of values for collision severity implicitly influences the behaviour of the autonomous vehicle and it is a highly ethical question (and a dilemma) to decide on the values, especially if humans are involved in the collision scenario.

At the moment, there are no exact rules for defining collision severity values. For purpose of illustration and simulation, we propose two groups of values in Table 1. In general, pedestrians are assigned the highest values. Buses have relatively lower values compared to pedestrians but still higher than cars. Bus stations and buildings are assigned the lowest values. As mentioned earlier, these values require thorough discussion and evaluation before their implementation in a real system.

Table 1: Collision Severity values of various objects.

| Objects | Setting 1 | Setting 2 |
|---|---|---|
| pedestrians | 40 | 200 |
| buses | 30 | 30 |
| cars | 20 | 20 |
| bus stations | 10 | 10 |
| buildings | 10 | 10 |

## 3 MINIMUM COLLISION SEVERITY PATH PLANNING

After getting the collision severity map, the path planning problem is designed. In this section, the vehicle model and the optimal control problem are introduced. We would like to emphasize that we are particularly interested in crucial scenarios where collisions are likely or cannot be avoided at all. The time to collision of such scenarios is typically short, in the range of 1-2 seconds, making pure braking impractical.

### 3.1 Vehicle Model

A kinematic vehicle model is used in this work and the equation of motion is as follows:

$$\dot{x} = vcos\varphi \quad (7)$$
$$\dot{y} = vsin\varphi \quad (8)$$
$$\dot{\varphi} = vtan\delta/L \quad (9)$$
$$\dot{v} = a \quad (10)$$

$$\dot{\delta} = (\delta_s - \delta)/\Delta T \quad (11)$$

Here, $x$ and $y$ are the coordinates of the car´s reference point (midpoint of rear axle) and $v$ denotes the velocity of the vehicle. The yaw angle is $\psi$ and the steering angle is $\delta$. $L$ represents the distance from the front axle to the rear axle of the vehicle, where the control input $a$ represents the acceleration of the vehicle. $\delta_s$ represents the desired steering angle, which is a control input. The constant value $\Delta T$ is used to model latency in the steering device.

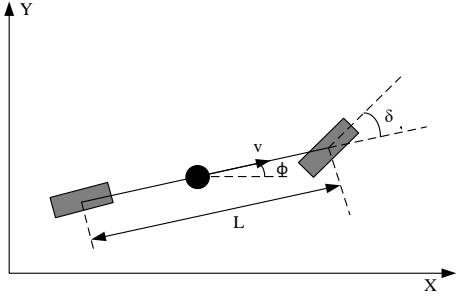

Figure 3: Vehicle model.

## 3.2 Optimal Control Problem

In section 2, we analyse the collision severity function, which we will use to formulate optimal control problems for path planning. Our first step is to solve the following optimal control problem (OCP$_1$), which aims at minimizing the total collision severity.

Minimize

$$J_1 = \sum_{i=1}^{N} \int_{t_0}^{t_f} cs(t, x(t), y(t), v(t))^2 dt \quad (12)$$

subject to system equations (7) to (11) and control constraints

$$\begin{cases} a_{min} \leq a \leq a_{max} \\ \delta_{min} \leq \delta_s \leq \delta_{max} \end{cases} \quad (13)$$

with initial conditions

$$(x(0), y(0), \varphi(0), v(0), \delta(0)) = (x_0, y_0, \varphi_0, v_0, \delta_0). \quad (14)$$

Here, $t_0$ and $t_f$ denote the initial and final time, respectively. N is the number of obstacles in the scenario. By minimizing the collision severity measure $J_1$, we obtain a minimal collision severity path. Let $J_1^*$ denote the optimal total collision severity value of (OCP$_1$). In general, the solutions are not unique and several paths with minimal collision severity measure exist. Among all such paths with minimal collision severity, we attempt to identify a suitable path for the autonomous vehicle.

To this end, in a second step, we solve the following optimal problem (OCP$_2$), which aims at finding a trajectory with minimized steering efforts among all trajectories with the same minimum total collision severity:

Minimize

$$J_2 = \int_{t_0}^{t_f} \delta_s(t)^2 dt \quad (15)$$

subject to system equations (7) to (11), constraints for acceleration and steering angle in (13), the initial conditions (14), and an additional collision severity constraint

$$\sum_{i=1}^{N} \int_{t_0}^{t_f} cs(t, x(t), y(t), v(t))^2 dt \leq J_1^* + \varepsilon. \quad (16)$$

Herein, $\varepsilon \geq 0$ is a small value, which can be used to slightly relax the optimal risk measure $J_1^*$ from OCP$_1$. This relaxation might enhance numerical stability.

## 4 SIMULATION RESULTS

In this section, the designed path planning method will be implemented in simulation software OCPID-DAE1 (Optimal Control and Parameter Identification with Differential Algebraic Equations of Index 1) for validation. OCPID-DAE1 is an optimal control tool that implements a direct shooting method. The resulting nonlinear optimization problem is solved by a sequential quadratic programming (SQP) method. Detailed information can be found in the book [18]. The intersection scenario is chosen for investigation, due to its complicated environment and high potential for accidents. At intersections: cars, buses, pedestrians and trees along the road boundaries are considered. Critical scenarios, where congestion or collisions cannot be avoided, are taken into account.

To further validate the effectiveness of the path planning controller, we also visualize the scenario using a popular and powerful game engine developed by Epic Games, Unreal Engine.

Scenarios of intersections are modelled, featuring moving cars, static cars, pedestrians, buses, bus stations, and buildings. Two scenarios, as shown in Figure 4 and 5, are studied. When encountering obstacles, braking is typically the first choice of drivers. However, there are situations, in which emergency braking may not be effective or the optimal choice. Therefore, in this paper, we aim to explore the potential of steering.

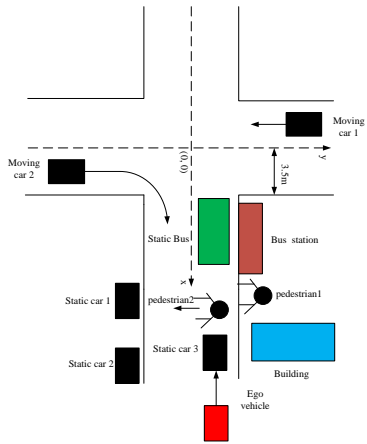

Figure 4: Scenario 1.

Table 2: Layout information of scenario 1.

| Objects | Coordinate | Velocity(m/s) |
|---|---|---|
| ego vehicle | (50, 1.75) | 10 |
| static car 1 | (21, -5) | 0 |
| static car 2 | (26, -5) | 0 |
| static car 3 | (30, 1.75) | 0 |
| bus | (16, 1.75) | 0 |
| pedestrian 1 | (20, 3.5) | 0 |
| pedestrian 2 | (24, 3.5) | 1 |
| moving car 1 | (-1.75, 18.5) | 10 |
| moving car 2 | (1.75, -18.5) | 10 |

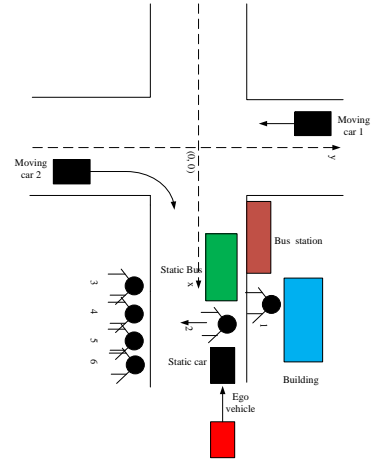

Figure 5: Scenario 2.

Table 3: Layout information of scenario 2.

| Objects | Coordinate | Velocity(m/s) |
|---|---|---|
| ego vehicle | (50, 1.75) | 10 |
| static car | (30, 1.75) | 0 |
| bus | (16, 1.75) | 0 |
| pedestrian 1 | (20, 3.5) | 0 |
| pedestrian 2 | (24, 3.5) | 1 |
| pedestrian 3 | (23, -5) | 0 |
| pedestrian 4 | (24, -5) | 0 |
| pedestrian 5 | (25, -5) | 0 |
| pedestrian 6 | (26, -5) | 0 |
| moving car 1 | (-1.75, 18.5) | 10 |
| moving car 2 | (1.75, -18.5) | 10 |

## 4.1 Case Study 1

In our algorithm, collision severity values of different objects are defined to adjust the vehicle's behaviour. In the first step, we aim to investigate how these values affect performance. Compared simulation of scenario 1 and 2 is presented. The layouts of scenario 1 and 2 are nearly identical. The only difference is that in scenario 1, there are two parking cars on the left side of the street. While in scenario 2, there are four pedestrians. The time to collision is around 2s in both scenarios.

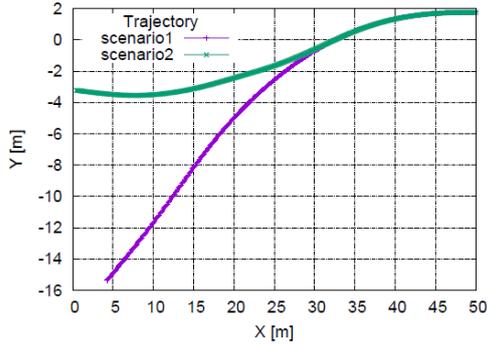

Figure 6: Trajectories of scenario 1 and 2.

The generated trajectories are displayed in Figure 6. The difference lies in the deviation from static cars and pedestrians. In scenario 1, the ego vehicle turns to collide with the static car 1. In scenario 2, the vehicle maintains a further distance from the pedestrians due to the higher collision severity value. This difference can be clearly observed in the visualizations of scenario 1 and 2, as shown in Figure 7 and 8.

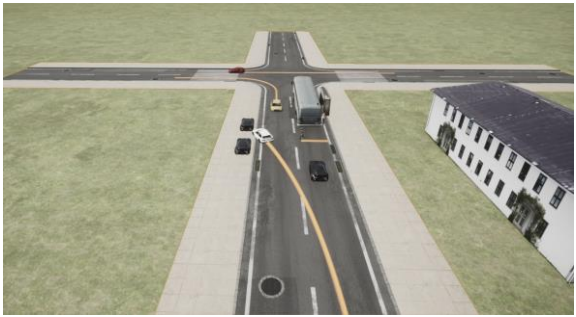

Figure 7: Visualization of Scenario 1.

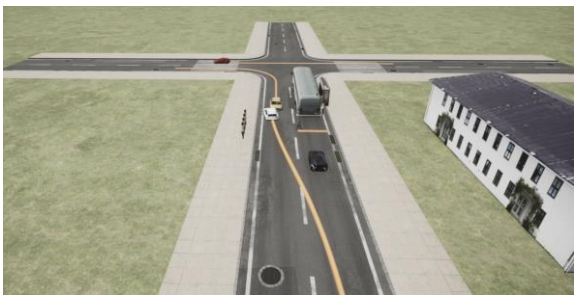

Figure 8: Visualization of Scenario 2.

The collision severity values in both scenarios are illustrated in Figure 9. It is evident that in scenario 2, the value is higher. This has an impact on the steering performance, as shown in Figure 10. It can be observed that in scenario 2, the steering angle returns to its original position earlier than in scenario 1. In both cases, the vehicle maintains a constant velocity. The velocity and acceleration are depicted in Figure 11 and Figure 12, respectively.

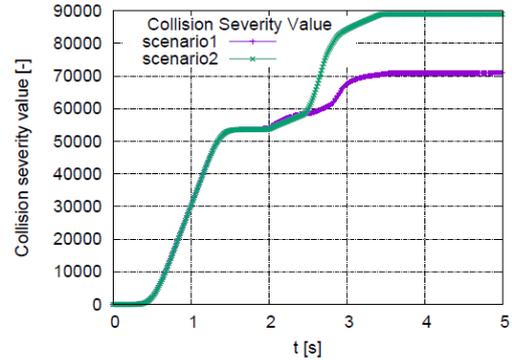

Figure 9: Collision severity values in scenario 1 and 2.

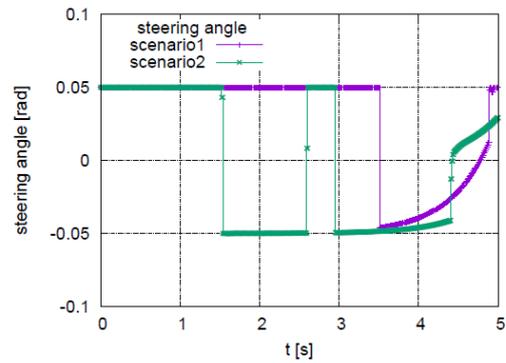

Figure 10: Steering angles of scenario 1 and 2.

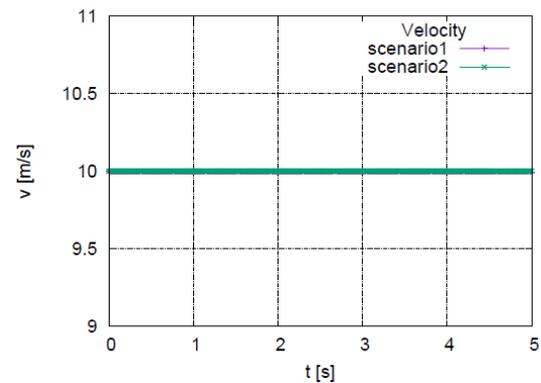

Figure 11: Velocities of scenario 1 and 2.

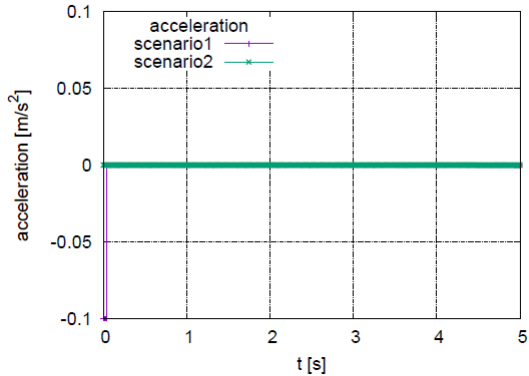

Figure 12: Accelerations of scenario 1 and 2.

## 4.2 Case Study 2

For autonomous driving, a frequent ethical question is how to react when a collision cannot be avoided, especially when facing different groups of people. Should the vehicle collide with one person or a group of person? Should it prioritize children or adults? In this section, we conduct simulations to explore a technical solution. In scenario 2, two pedestrians approach from one side and on the other side are four pedestrians. Two conditions of scenario 2 are modelled. In condition 1, the collision severity values for all 6 pedestrians are the same. In condition 2, it is assumed that pedestrian 2 is a small child, who is going to traverse the street. Therefore, a higher risk value of 200 is assigned to pedestrian 2, while others maintain the same value of 40. Simulation results are analysed.

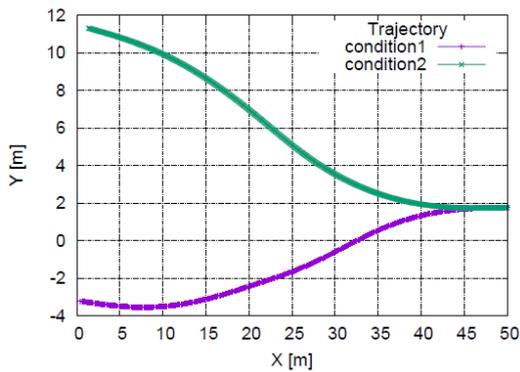

Figure 13: Trajectories of condition 1 and 2.

The trajectories depicted in Figure 13 reveal distinct behaviours of the vehicle under two conditions. In condition 2, attributed to the higher collision severity value of pedestrian 2, the vehicle navigates directly around them from the opposite side. This difference can be evidently observed in the visualizations of condition 1 and 2, as shown in Figure 14 and 15.

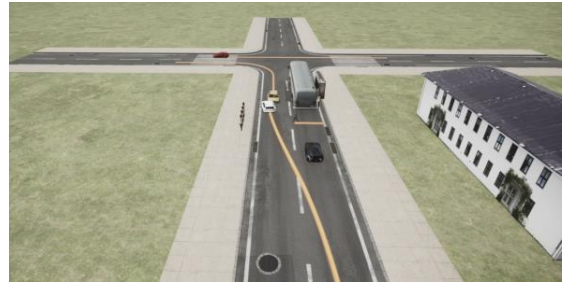

Figure 14: Visualization of condition 1.

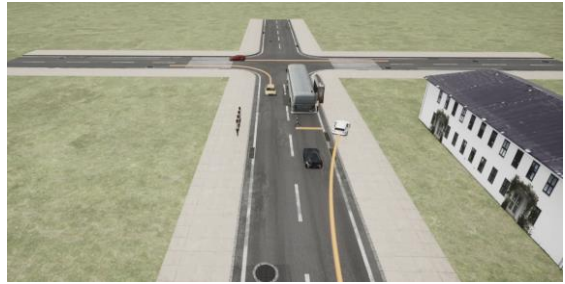

Figure 15: Visualization of condition 2.

In both of these conditions, we aim to observe the steering behaviour, and therefore, constraints are imposed on accelerations. Collision severity values are outlined in Figure 16, revealing that Condition 2 exhibits larger values compared to Condition 1. The steering angles in both conditions differ initially, highlighting how collision severity values directly impact the car's behaviour. This insight suggests a potential solution to address the identified issue.

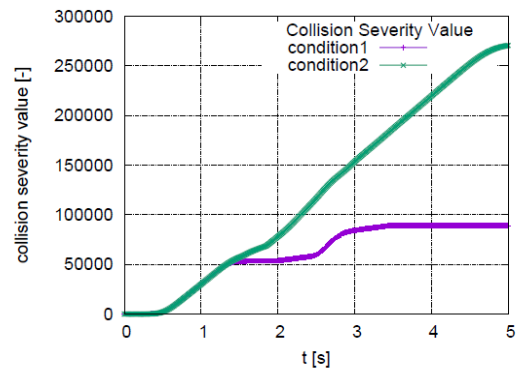

Figure 16: Collision severity value of condition 1 and 2.

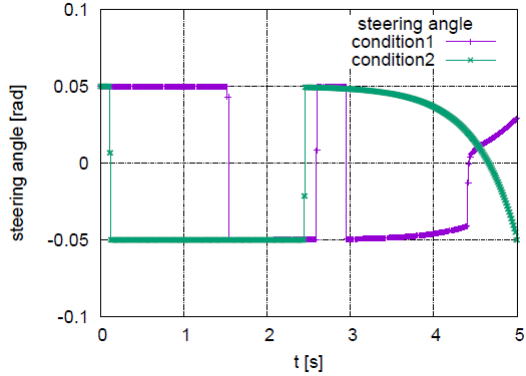

Figure 17: Steering angles of condition 1 and 2.

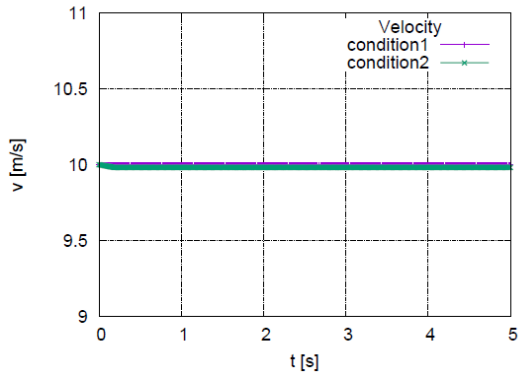

Figure 18: Velocities of condition 1 and 2.

The velocity and acceleration profiles in Figure 18 and 19 indicate that the vehicle maintains an almost constant speed.

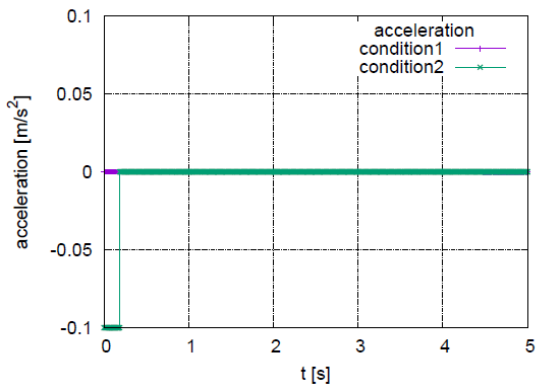

Figure 19: Accelerations of condition 1 and 2.

## 5 CONCLUSIONS

This paper presents the development of a collision severity map encompassing both static and dynamic obstacles. The evaluation of collision severity values of different objects is explored, which is crucial for decision-making, particularly in critical situations. Collision severity value and steering efforts are minimized to obtain an optimal path. Simulation results demonstrate that the method can serve as a technical solution for addressing challenges in autonomous driving, including collisions with various objects and ethical dilemmas. For future work, further investigation into the influence of different collision severity values on the vehicle performance is an interesting topic. Additionally, the prospective implementation of this algorithm in a real vehicle is also a promising direction.

## ACKNOWLEDGEMENTS

This work is funded by detc.bw- Digitalization and Technology Research Center of the Bundeswehr München [project EMERGENCY-VRD. MORE]. detc.bw is funded by the European Union- Next Generation EU.